\documentclass[lettersize,journal]{IEEEtran}
\usepackage{amsmath,amssymb,amsfonts}
\usepackage{algorithmic}
\usepackage{graphicx}
\usepackage{subcaption}
\usepackage{textcomp}
\usepackage[table]{xcolor}
\usepackage{multirow}
\usepackage[misc]{ifsym}
\usepackage{url} 
\usepackage{booktabs}
\usepackage{bm}
\usepackage[numbers,sort&compress]{natbib}
\usepackage{tikz}
\usepackage{circledsteps}
\usepackage{listings}
\usepackage{soul}
\usepackage{xspace}
\usepackage[autostyle]{csquotes}  
\usepackage{todonotes}
\usepackage{pgfplots}
\usepackage{pgfplotstable}
\usepackage{threeparttable}
\usepackage{tabularx}
\usepackage{hyperref}

\usepackage{pifont}
\usepackage[T1]{fontenc}
\usepackage[normalem]{ulem}
\usepackage[most]{tcolorbox}
\usepackage{newfloat}
\usepackage{adjustbox}
\usepackage{enumitem}
\usepackage{colortbl} 

\setuptodonotes{inline}

\definecolor{coolblack}{rgb}{0.0, 0.18, 0.39}

\definecolor{formalshade}{rgb}{1.0,1.0,1.0}
\definecolor{side}{rgb}{0.0,0.2,0.6}

\definecolor{grayrow}{rgb}{0.7,0.7,0.7}
\definecolor{my4grey}{HTML}{525252}
\definecolor{my3grey}{HTML}{969696}
\definecolor{my2grey}{HTML}{cccccc}
\definecolor{my1grey}{HTML}{f7f7f7}

\usepackage{amsthm}

\usepackage[strict]{changepage}
  \definecolor{ABlue}{HTML}{127bca}
 \definecolor{LHScolor}{HTML}{555555}
\usepackage{framed}

\definecolor{formalshade}{rgb}{1.0,1.0,1.0}
\definecolor{side}{rgb}{0.0,0.2,0.6}

\def\BibTeX{{\rm B\kern-.05em{\sc i\kern-.025em b}\kern-.08em
    T\kern-.1667em\lower.7ex\hbox{E}\kern-.125emX}}

\begin{document}

\title{Evaluating Fuzz Testing for Reinforcement Learning Agents}

\author{Zhibin Kang, Hanmo You, Dong Wang, Haiming Zheng, Junjie Chen*
\IEEEcompsocitemizethanks{\IEEEcompsocthanksitem{Z. Kang, H. You, D. Wang, H. Zheng, J. Chen are with the College of Intelligence and Computing, Tianjin University, Tianjin, China.

E-mail: \{kzb, youhanmo, dong\_w, zhenghm, junjiechen\}@tju.edu.cn}
\IEEEcompsocthanksitem Corresponding author: Junjie Chen.



}
}

\markboth{Journal of \LaTeX\ Class Files,~Vol.~18, No.~9, September~2020}%
{How to Use the IEEEtran \LaTeX \ Templates}

\maketitle

\begin{abstract}
Reinforcement Learning (RL) agents are increasingly deployed in safety-critical domains such as robotics, autonomous driving, and drone control, where unexpected behaviors may lead to severe real-world consequences. 
Fuzz testing has recently emerged as a promising method for exploring the vast state spaces of RL agents and exposing crashes.
Although numerous RL fuzzing methods have been proposed, existing studies often differ in evaluation settings, baselines, and metrics, making it difficult to draw reliable conclusions about their relative effectiveness and practical usefulness.
To address this gap, we present the first comprehensive empirical study that systematically evaluates RL fuzzing methods from four complementary perspectives: effectiveness, diversity, efficiency, and practical utility. 
We benchmark five state-of-the-art methods alongside random testing under unified configurations across three environments of increasing complexity (MountainCar, BipedalWalker, and CARLA), and further assess the downstream usefulness of detected crashes for agent robustness improvement and safety monitoring.
Our results reveal several key insights.
For instance,throughput-oriented methods like MDPFuzz demonstrate superior effectiveness and efficiency in crash discovery, while methods explicitly designed to encourage exploration like SeqDivFuzz excel at uncovering diverse crash behaviors. 
We also show that fuzzing-generated crashes can meaningfully improve agent robustness and enable accurate safety monitoring with strong cross-method generalization.
Beyond these empirical findings, we distill actionable guidance for both researchers and practitioners, highlighting the benefits of combining complementary fuzzing strategies and adopting multi-level diversity analysis to achieve more comprehensive and practical RL testing.
\end{abstract}


\maketitle

\section{Introduction}
\label{sec:introduction}
Reinforcement Learning (RL) agents have been increasingly deployed in domains that are closely intertwined with human life, including safety-critical scenarios such as drone navigation~\cite{DBLP:journals/nca/HodgeHA21}, robotic control~\cite{DBLP:conf/icra/JohanninkBNLKLO19}, and autonomous driving~\cite{DBLP:journals/aei/ZhouHLLCS25}. 
An RL agent typically learns decision-making policies through repeated interactions with an environment, receiving feedback in the form of rewards and penalties, and continuously adapting its behavior to maximize long-term objectives.~\cite{o2017learning}
Similar to traditional software systems, they may exhibit unexpected or erroneous behaviors, particularly when exposed to rare, complex, or previously unseen environmental conditions~\cite{DBLP:conf/icse/ThomasBHWJRT25}. The potential risks of such crashes have been underscored by real-world incidents involving learning-based autonomous systems, such as the widely reported Cruise autonomous vehicle pedestrian collision~\cite{Cruise}, which revealed the challenges of achieving reliable decision-making in highly dynamic and uncertain environments.
These observations highlight the practical urgency and societal importance of reliable RL agents.

Testing is a critical means of ensuring the quality and reliability of RL agents ~\cite{mdpfuzz,mdpfuzznew}.
Among the various testing methods, fuzzing, which generates large volumes of diverse and potentially adversarial test inputs to trigger unexpected behaviors, has proven particularly effective in exploring the vast and complex state spaces of RL agents~\cite{survey, wang2025survey}. 
As a result, fuzzing has emerged as a prominent research direction for exposing potential crashes, and a number of fuzzing-based methods have been proposed in recent years. 
For example, Pang et al.~\cite{mdpfuzz} introduced MDPFuzz, which employs Gaussian Mixture Models (GMMs) to guide the generation of test inputs and increase the likelihood of triggering crashes. Li et al.~\cite{g-model} proposed a generation-based method, which leverages diffusion models to learn the distribution of normal test inputs and subsequently generate diverse abnormal test inputs that deviate from this learned distribution. These methods have demonstrated promising effectiveness and have laid a solid foundation for subsequent research on RL agent fuzz testing.

While these methods have shown encouraging progress, selecting an appropriate RL fuzzing method for real-world deployment remains challenging. Existing methods differ substantially in their design objectives and optimization strategies, making their strengths and limitations difficult to assess from individual studies alone. Consequently, practitioners and researchers lack reliable empirical evidence for choosing appropriate methods under different testing requirements~\cite{DBLP:conf/icse/YouWLC25}. However, current studies~\cite{mdpfuzznew,seqfuzz,g-model,curefuzz,QDfuzz} provide limited support for such decision making due to several methodological shortcomings.
\textbf{First, comparative evaluation remains insufficient.} 
Many proposed methods are evaluated in isolation and are not systematically included as baselines in subsequent studies, limiting the ability to establish clear performance improvements over prior work.
Even fundamental strategies such as random testing, which provide an essential stochastic lower bound, are frequently omitted from comparisons.
Moreover, differences in implementation versions (e.g., early version ~\cite{mdpfuzz} versus updated releases of MDPFuzz~\cite{mdpfuzznew}) may influence experimental outcomes, potentially introducing variability in reported results.
\textbf{Second, comparative experiments often lack fairness and standardization.} 
Different works adopt inconsistent parameter settings, heterogeneous evaluation metrics, and varying benchmark tasks, making direct comparisons difficult.
For example, certain methods~\cite{QDfuzz,smarla} fail to investigate complex benchmark scenarios, while others~\cite{mdpfuzz} do not incorporate diversity as a key metric in their evaluation.
Such inconsistencies obscure the true effectiveness and efficiency of competing fuzzing methods and weaken the validity of cross-paper conclusions.
\textbf{Third, attention to practical usefulness remains limited.} 
Existing works primarily focus on maximizing crash discovery, while paying less attention to how detected crashes contribute to downstream benefits.
For example, some methods~\cite{QDfuzz,seqfuzz} emphasize crash diversity but do not examine whether the discovered crashes can enhance agent robustness through retraining or stress testing.
Similarly, safety monitoring~\cite{DRLFailureMonitor}, i.e., leveraging detected risky states to predict or warn about crash episodes in advance, remains underexplored, despite its importance for real-world deployment. 
Consequently, practitioners cannot determine whether improvements in crash discovery translate into greater engineering benefits for downstream software assurance tasks.

To address the aforementioned gaps, we conduct the first comprehensive empirical study to systematically evaluate the effectiveness and efficiency of existing RL fuzzing methods.
Specifically: 
1) We include five state-of-the-art fuzzing methods~\cite{mdpfuzznew,seqfuzz,g-model,curefuzz, QDfuzz}, together with the fundamental random testing strategy, and evaluate their updated and corrected implementations to ensure a sufficiently thorough and reliable assessment.
2) We adopt unified experimental configurations and evaluate all methods on a diverse set of benchmarks spanning three scale levels (i.e., MountainCar, BipedalWalker, and CARLA), enabling a fair and controlled comparison.
3) We extend the evaluation beyond crash detection by considering two downstream tasks that leverage the detected crashes, agent robustness improvement and safety monitoring, to assess the practical utility of fuzzing in real-world deployment scenarios.
To structurally guide our empirical study, we formulate the following four research questions (RQs):

\textbf{- RQ1: How effective are the studied fuzzing methods in crash detection?}
Overall, methods that emphasize high testing throughput tend to achieve stronger crash discovery performance, with MDPFuzz consistently outperforming other methods across benchmarks. 
For relatively simple tasks, lightweight strategies remain competitive (i.e., random testing), whereas for more complex environments, guided exploration mechanisms become increasingly important for effective crash discovery. 


\textbf{- RQ2: How diverse are the crashes detected by the studied fuzzing methods?}
Methods explicitly designed to encourage exploration, such as SeqDivFuzz, tend to uncover more diverse crash behaviors, whereas efficiency-oriented methods like MDPFuzz prioritize rapid crash discovery but exhibit limited behavioral diversity. 
These findings highlight an inherent trade-off between crash quantity and crash diversity in RL fuzz testing.

\textbf{- RQ3: How efficient are the studied fuzzing methods?}
Results show that mutation-based fuzzing methods demonstrate substantially higher efficiency than generative methods, requiring far fewer iterations to trigger each unique crash. 
These results highlight that fast, guided exploration is critical for efficient RL fuzz testing.

\textbf{- RQ4: How useful are the crashes detected by the studied fuzzing methods for RL agents?}
Fuzzing-generated crashes demonstrate practical value for downstream tasks. 
When used for agent repair, most methods improve robustness, with diversity-oriented methods achieving the largest gains. For safety monitoring, models trained on crash-triggering inputs achieve strong detection performance, typically exceeding 95\% accuracy even in cross-method settings, indicating that fuzzing exposes both method-specific and shared crash characteristics that generalize well in practice.

Beyond quantitative comparisons, our study yields several practical implications for the design and evaluation of RL fuzz testing.
We observe that random testing is frequently overlooked and rarely adopted as a baseline. It delivers better crash detection effectiveness, diversity, and efficiency than many proposed methods, particularly for simple tasks. This delivers an important reminder for researchers.
Different fuzzers exhibit complementary exploration behaviors, suggesting that combining multiple strategies and adopting multi-level diversity metrics could improve crash coverage.
Our results further highlight the importance of standardized evaluation protocols, scalable and task-agnostic designs, and testing under realistic deployment scenarios rather than simplified toy environments.

\smallskip
\noindent
\textbf{Contributions.} To sum up, our work makes the following major contributions: (i) We conduct the first large-scale systematic empirical study of RL fuzz testing, benchmarking six representative methods under unified configurations across environments of varying complexity.
Our evaluation spans four comprehensive dimensions, including effectiveness, diversity, efficiency, and practical utility. 
(ii) Through controlled and in-depth analysis, we reveal the strengths, limitations, and trade-offs of existing fuzzing strategies. We further demonstrate the downstream value of fuzzing and distil a set of actionable guidelines to inform both practitioners and researchers.
(iii) We re-implement and standardize the compared fuzz methods, construct an evaluation pipeline to facilitate fair and reproducible comparison, and publicly release all used code, datasets, and experimental artifacts to support future studies~\cite{EMD2024}.
\section{Background and Related Work}
\label{sec:background}
\subsection{Reinforcement Learning}

\begin{figure}[t]
    \centering
    \includegraphics[width=0.8\linewidth]{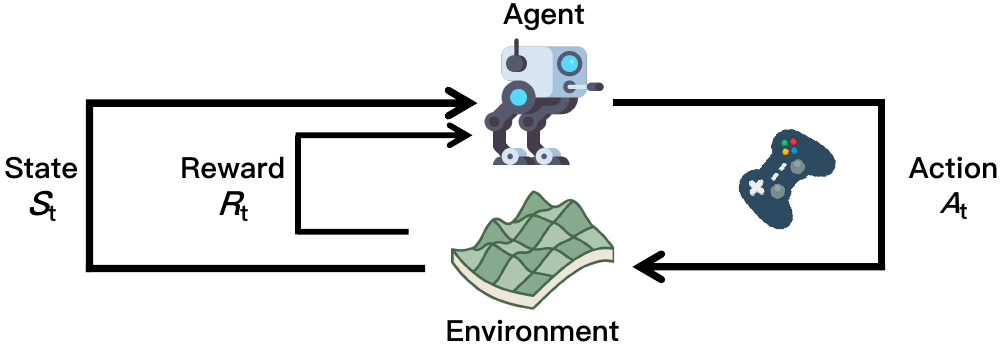}
    \caption{The General Framework of RL}
    \label{fig:rl}
\end{figure}

RL provides a principled mathematical framework to describe how an agent learns to perform a specific task by reinforcing its rewarding behaviors~\cite{barto2021reinforcement,wang2025survey}. RL is different from standard machine learning, which typically operates with independently and identically distributed samples that contain explicit ground truth labels. RL, by contrast, focuses on the interaction between actions and the rewards they generate. The general framework of RL is shown in Figure~\ref{fig:rl}. At every timestep $t$, an agent recognizes the state $s_t$ it is in and evaluates the desirability of that
state, known as the reward $r_t$. Based on this information, the agent draws an action $a_t$ to cause the environment to evolve
into the next state, and the loop continues. The core goal of the agent is to learn an optimal policy distribution $\pi$. This policy specifies the way to choose actions under a given state, so as to maximize the cumulative rewards over time.

Formally, RL problems are formulated based on the Markov Decision Process (MDP)~\cite{puterman2014markov}. MDP is typically represented as a tuple $(\mathcal{S},\mathcal{A},\mathcal{P},\mathcal{R},\gamma)$. In this tuple, $\mathcal{S}$ denotes the state space. It corresponds to all possible observations that the agent can perceive at each step of execution. $\mathcal{A}$ represents the set of possible actions. These actions can be either continuous or discrete, and they are the outputs that the agent returns to the environment. $\mathcal{P}$ stands for the transition probability. Specifically, it refers to the probability of transitioning to the next state when a specific action is taken in the current state. $\mathcal{R}$ is the reward function. Rewards serve as the feedback that the agent receives after executing each action. $\gamma$ is a discount factor, which is used to weight the importance of immediate and future rewards. 

\subsection{Testing Reinforcement Learning Agents}

Testing plays a critical role in ensuring the quality and reliability of RL agents, as it aims to generate test inputs that can expose faulty or unsafe behaviors during interaction with the environment~\cite{DBLP:journals/tosem/BiagiolaT22}. 
In the context of RL, a test input typically corresponds to a specific environment configuration or initial state from which the decision-making process begins. 
The agent under test (i.e., the agent) then interacts with the environment step-by-step, producing actions and receiving feedback, thereby forming an execution trajectory. By analyzing the outcomes of these trajectories (e.g., crashes, violations, and abnormal rewards), testing can effectively reveal weaknesses in the learned policy.

There are various methods for testing RL agents, with adversarial testing being a typical one that often adopts a white-box strategy. Phan et al.~\cite{ARTS} proposed ARTS to generate adversarial inputs in multi-agent RL (MARL) systems. During testing, ARTS designates a portion of agents as antagonists according to a specified ratio, which creates varied crashes in each episode. Guo et al. ~\cite{MARLSafe} introduced MARLSafe, a method that applies state perturbations through gradient-based FGSM to mislead the agents’ decisions by reducing the probability of optimal actions. Zhou et al. ~\cite{DE} employed differential evolution to generate worst-case joint actions, focusing on those agents whose actions have the most significant impact on team coordination. A common drawback of these adversarial testing methods is that they are usually time-consuming due to their reliance on white-box information.

Another important testing method is fuzzing, a widely used method in software testing that automates the generation of inputs to identify vulnerabilities, crashes, or other unexpected behaviors in software programs~\cite{DBLP:journals/csur/ZhuWCX22}. This method has also been proven effective and widely applied in testing RL agents. 
Ul Haq et al.~\cite{DBLP:conf/icse/HaqSB23} proposed MORLOT, a hybrid framework that integrates RL with many-objective search to test DNN-enabled systems under dynamic environments and multiple requirements.
Zolfagharian et al.~\cite{starla} presented STARLA, a data-box method that uses machine learning models to predict episode-level faults and steer the search toward failure-prone trajectories. 
Pang et al.~\cite{mdpfuzz} used GMMs to guide coverage-based fuzzing, aiming to discover more crashes in RL agents. Mazouni et al. replicated this GMM-guided method and extended its evaluation to additional RL tasks. 
Wang et al.~\cite{seqfuzz} proposed SeqDivFuzz, an optimized fuzzing framework that boosts testing effectiveness and efficiency for RL agents by integrating a sequence diversity inference module to early terminate non-diverse test cases during MDP interaction.
He et al.~\cite{curefuzz} proposed CureFuzz, a curiosity-driven black-box fuzzing framework for RL that improves the detection of diverse crash-triggering crashes by leveraging a curiosity mechanism to measure novelty and a multi-objective seed selection method to balance exploration and crash generation.
Li et al.~\cite{AgentFuzz} proposed AgentFuzz, a method that generates diverse initial states, monitors the total reward of each episode, and labels episodes with rewards below a preset threshold as failed. 
Mazouni et al.~\cite{QDfuzz} proposed a framework that boosts crash diversity in RL testing using quality diversity optimization to explore diverse crash-triggering inputs.
Li et al.~\cite{g-model} proposed a generation-based RL fuzzing framework that boosts crash detection and behavior diversity by leveraging a generative diffusion model for test input generation.
Ma et al.~\cite{MASTest} proposed MASTest, a framework that enhances test diversity in MARL by using grid-based state abstraction to simplify complexity and compare agent behaviors. 

Compared with adversarial testing, fuzzing usually uses black-box and grey-box information for testing, making it easier to generalize to different RL agents and scenarios. 
As highlighted in the recent survey~\cite{survey}, its strong effectiveness and broad applicability have made fuzzing a major focus of current RL testing research; therefore, we target it as the focus of our main investigation.
\section{Study Design}
\label{sec:evaluation_design}

Using three widely adopted benchmarks spanning different scales (i.e.,  MountainCar, BipedalWalker, and CARLA), we first evaluate the effectiveness of existing methods in discovering crashes in RL agents through fuzz testing (\textbf{RQ1}).
Next, we analyze the diversity of crash-triggering test inputs generated by each fuzzing method (\textbf{RQ2}).
We then measure the efficiency of each method in terms of time cost and resource consumption (\textbf{RQ3}).
Finally, we incorporate two downstream tasks to evaluate the practical usefulness of the detected crashes, examining their impact on agent robustness improvement and safety monitoring (\textbf{RQ4}).

\subsection{Studied Benchmark}
This study aims to conduct a comprehensive and unified comparison of state-of-the-art methods. Specifically, we select the benchmarks based on the following criteria: (I) widely used in existing fuzzing literature; (II) coverage of diverse application tasks (e.g., robotic control and autonomous driving), agent scales (ranging from 135k to 1 million weight parameters), and representative RL algorithms.



To ensure all fuzzing methods are evaluated under conditions that closely resemble real-world deployment, we select robust RL agents as the subjects under test rather than agents that fail easily under random testing. Specifically, following the methodology of Uesato et al.~\cite{DBLP:conf/iclr/UesatoKSERADHK19}, we first generate 1,000 random configurations for each RL agent using a random testing strategy and estimate its failure rate to assess its robustness. We then select the RL agents with relatively low failure rates reported in prior studies~\cite{rl-zoo3,roach} for subsequent experiments, thereby avoiding the situation where excessive intrinsic vulnerabilities of the RL agent obscure the performance differences among competing fuzzing methods. Furthermore, the same set of 1,000 randomly generated configurations is retained as an independent test set to evaluate the generalization performance of RL agents after downstream tasks, such as robustness improvement, and to analyze the proportion of regression faults introduced during these processes.

\textbf{MountainCar.}~\cite{gym}
It serves as a low-dimensional control benchmark for RL. This task requires an underpowered car to accumulate momentum through back-and-forth swings and climb from the bottom of the valley to the top.
The initial seed input is a 2-dimensional vector, including the initial height and initial velocity of the car.
We follow the experimental settings from existing work~\cite{smarla,starla}:
We adopt Deep Q-Network as the RL algorithm with 134k trainable weight parameters with 1.1\% failure rate. Following existing work~\cite{smarla,starla}, we use 10,000 initial seed inputs and only mutate the initial position while fixing the initial velocity to avoid invalid test inputs such as starting at the valley bottom with zero velocity. A test case is regarded as successful if the car’s position exceeds 0.5 within 200 steps; otherwise, it is labeled as a crash.

\textbf{BipedalWalker.}~\cite{gym}
It is a high-difficulty continuous control task under the OpenAI Gym~\cite{gym} framework. This task requires controlling a bipedal robot to maintain balance and move forward efficiently on generated rough terrain. The initial seed input is a 15-dimensional continuous state vector, including information about the environmental terrain. We follow the complete experimental settings of existing work~\cite{mdpfuzz,mdpfuzznew, QDfuzz,seqfuzz,curefuzz,g-model}:
We adopt Truncated Quantile Critics as the RL algorithm with 690k trainable weight parameters with 1.6\% failure rate. Following existing work~\cite{CaRL,rl-zoo3}, we use 1,000 initial seed inputs and perform mutation operations on all 15 dimensions. A test case is labeled as a crash if the robot falls over within 300 steps. 

\textbf{CARLA.}~\cite{carla}
It is a high-fidelity open-source simulator for autonomous driving. It provides a realistic urban driving environment with complex traffic flow and is often used to detect safety vulnerabilities of autonomous driving systems in complex interactive scenarios. The initial seed test input is a 66-dimensional vector that defines the test input. It includes the start and end points of the route, weather combinations, the initial state of the ego-vehicle, and the positions of NPC vehicles. We follow the experimental settings from existing work~\cite{mdpfuzz,mdpfuzznew,QDfuzz,seqfuzz,curefuzz,g-model}:
We adopt the state-of-the-art Roach agent~\cite{CaRL, roach} as the RL algorithm, which is trained via  Proximal Policy Optimization with 1.5 million trainable weight parameters, with 7.2\% failure rate.
Following existing work~\cite{mdpfuzz,mdpfuzznew,seqfuzz,curefuzz,g-model}, we use 100 initial seed inputs and perform mutation operations on 63 dimensions of the state vector.
A test input is regarded as a crash if the vehicle collides or fails to reach the destination within 200 steps.

\subsection{Validity Checks}
\label{subsec:validity-checks}



Ensuring the validity of generated test inputs is essential for the fair evaluation of RL fuzzing methods. Specifically, a valid test input should correspond to a feasible environment configuration in which the target task is theoretically achievable, rather than an inherently infeasible test input caused by violating implicit environmental constraints. Otherwise, invalid test inputs may distort the assessment of an RL agent's robustness and lead to an overestimation of crash rates.
However, directly verifying the validity of generated test inputs is challenging in complex RL environments. For example, in CARLA, each test input is specified by a 66-dimensional environment configuration whose feasibility depends on numerous implicit constraints, making it impractical to determine the validity of every generated configuration individually.

To address this challenge, we adopt a conservative \textit{differential testing} strategy for validity checking. Specifically, for each agent under test, we employ an independently trained, highly robust validation policy. A generated test input is considered valid only if the validation policy successfully completes the task while the target agent fails~\cite{DBLP:journals/cacm/PeiCYJ19}. In this case, the observed crash is attributed to deficiencies of the target agent rather than to an infeasible environment configuration.

For the three benchmark environments, we select validation policies with the lowest failure rates under random testing. Specifically, we use a PPO policy for MountainCar, an SAC policy for BipedalWalker, and a CaRL policy for CARLA. Their failure rates under random testing are 0\%, 0\%, and 2.5\%, respectively~\cite{rl-zoo3,CaRL}. Although the failure rate of the CaRL policy is relatively higher, this reflects the substantially greater complexity of autonomous driving compared with classical control tasks. Moreover, it is the most robust policy available for the CARLA benchmark in our experiments, making it a reliable validation policy for differential testing.

\subsection{Studied Methods}

To ensure the validity and representativeness of our empirical evaluation, we rigorously selected baseline methods according to two criteria: (I) \textbf{widespread adoption in the research community}, and (II) \textbf{practical replicability}.
Specifically, we prioritized methods that have been extensively studied and cited in prior work, reflecting their relevance and influence within the RL fuzzing literature. 
At the same time, we required that each selected method provide accessible and well-documented source code or official replication packages, enabling reliable reproduction of experimental results. 
Methods that lacked publicly available implementations or posed significant replicability challenges ~\cite{AgentFuzz,DBLP:conf/ijcai/TapplerCAK22} were therefore excluded. 
We note that some fuzzing methods~\cite{MASTest} are designed specifically for certain types of RL algorithms (e.g., multi-agent reinforcement learning (MARL)~\cite{MASTest}) or focus on evaluating other system properties (e.g., plasticity or adaptation to new environments~\cite{DBLP:journals/tosem/BiagiolaT22,DBLP:journals/tosem/BiagiolaT24}). 
These methods often exploit characteristics unique to MARL settings (e.g., team diversity~\cite{MASTest}) or adopt evaluation metrics centered on performance degradation and recovery rather than crash detection.
As a result, their assumptions, objectives, and metrics are not directly comparable to fuzzing-based crash discovery in single-agent RL agents. Since our study specifically focuses on crash detection for single-agent RL, we do not include these methods and instead view their evaluation as an important direction for future work.
To the end, we include the following five representative or state-of-the-art fuzzing methods.
\begin{itemize}[leftmargin=0.4cm]
    \item \textbf{MDPFuzz}~\cite{mdpfuzz}: As a pioneering black-box fuzzing framework designed for Markov Decision Processes, this method employs GMMs to quantify coverage. Nevertheless, a subsequent replication study~\cite{mdpfuzznew} has demonstrated that simplified fuzzing heuristics often outperform their original coverage-guided mechanisms in terms of crash detection capability and efficiency. Therefore, we adopt its updated version for the purpose of evaluation.
    

    \item \textbf{CureFuzz}~\cite{curefuzz}:
    This method incorporates a curiosity-driven mechanism that measures scenario novelty using the prediction error between a target network and a predictor network. 
    The resulting error signal is used as feedback to guide test generation, enabling CureFuzz to move beyond traditional coverage metrics and more effectively uncover diverse and severe crash modes.

    \item \textbf{Generative Model-Based Testing}~\cite{g-model}: This framework integrates a generative diffusion model pre-trained on nominal trajectories. By fine-tuning with a "terminal state novelty" reward, it exploits the model's distributional fitting capabilities to synthesise diverse test cases that target long-tail risks. 
    We refer to this method as \textbf{G-Model}.

    \item \textbf{SeqDivFuzz}~\cite{seqfuzz}:
    This method addresses the test oracle efficiency problem by employing a Siamese Network–based evaluator to estimate sequence diversity at runtime. 
    By pruning redundant, low-diversity test cases early in the testing process, it more effectively allocates the computational budget, thereby enabling the discovery of a larger number of unique crashes.

    \item \textbf{QD-based Policy Testing}~\cite{QDfuzz}: This framework shifts the testing objective from merely detecting crashes to promoting crash diversity. 
    It applies Quality Diversity (QD) algorithms to model the testing process as a simultaneous search for high-performing and behaviorally distinct solutions within the behavioral space. We name it as \textbf{QDFuzz}. 
    
\end{itemize}
In addition, to address a common methodological gap in prior studies, we explicitly incorporated a \textbf{Random Testing} strategy as a baseline. 
By generating test inputs through uninformed, stochastic sampling without guidance from learned models or heuristics, Random Testing provides a principled lower bound on achievable performance.
This baseline enables a clearer interpretation of the effectiveness of advanced fuzzing methods by distinguishing genuine methodological gains from improvements that could arise from random exploration.

\subsection{Evaluation Metrics}
\label{subsec:metrics}

To comprehensively evaluate the performance of existing methods, we establish an evaluation framework spanning three complementary dimensions: Effectiveness, Diversity, and Efficiency, following commonly adopted practices in prior fuzzing and RL testing~\cite{mdpfuzz,mdpfuzznew, QDfuzz,seqfuzz,curefuzz,g-model}. \textbf{To guarantee practical plausibility, all generated test inputs undergo rigorous validity checks, and only valid cases are included in our subsequent analysis.}

\textbf {Effectiveness.}
Consistent with the standard evaluation paradigm in the fuzzing domain~\cite{vuzzer}, we use the number of valid \ul{\textit{ unique crashes} (\#UC)} as the primary effectiveness metric. 
In the RL fuzzing setting, a \textit{unique crash} is jointly determined by the initial seed (i.e., initial state) and the corresponding crash trajectory (i.e., behavior): two crashes are regarded as distinct if they differ in either the initial seed or the trajectory.
For example, an autonomous car may crash into the same obstacle from two slightly different initial positions, but the crashes stem from different model flaws~\cite{DBLP:conf/icse/YouWCLL23} (e.g., one from misjudgment and the other from slow response).
Specifically, under the same initial seed corpus and environment configurations, we record the total number of unique crashes detected by each fuzzing method within a fixed 12-hour time budget. A higher value indicates stronger crash detection capability.

\textbf {Diversity.}
While detecting a large number of crashes is important, repeatedly triggering similar crashes provides limited additional insight into system vulnerabilities. Therefore, we evaluate the diversity of detected crashes using two complementary metrics proposed by Bartlett et al.~\cite{bartlett2025pursuit}: \textit{Input Diversity} and \textit{Output Diversity}. Both metrics estimate diversity by clustering crash samples after PCA-based dimensionality reduction. Following the original study~\cite{bartlett2025pursuit}, we determine the optimal number of clusters using a silhouette coefficient improvement threshold of 20\%.

\ul{\textit{Input Diversity} (ID)} measures the breadth of exploration in the environmental configuration space. Specifically, it clusters vectorized test inputs and quantifies diversity by the number of clusters covered by crash-triggering test inputs. A higher ID indicates that crashes are triggered under more environmental conditions, suggesting broader coverage of the input space.

\ul{\textit{Output Diversity} (OD)} measures the diversity of crash behaviors in the execution space. Specifically, it clusters the execution trajectories of crash-triggering test inputs (e.g., position sequences and motion-state trajectories) and quantifies diversity by the number of distinct behavioral clusters represented by the detected crashes. A higher OD indicates that the fuzzing method exposes a wider variety of behavioral crash modes, rather than repeatedly triggering similar crashes.

\textbf {Efficiency.}
To assess the cost of crash discovery, we evaluate efficiency from both time and search perspectives using two complementary metrics.

\ul{\textit{UC Discovery AUC} (UD-AUC)}~\cite{efficiency} measures the temporal efficiency of discovering unique crash categories. Specifically, it is computed as the area under the cumulative unique crash discovery curve with respect to elapsed testing time. A higher UD-AUC indicates that a fuzzing method discovers more unique crashes earlier under a fixed testing budget.


\ul{\textit{Average Generations per Unique Crash} (AG/UC)}~\cite{generation} 
measures the average number of mutation operations required to trigger each crash. 
By focusing on iteration counts rather than elapsed time, this metric reduces the influence of hardware differences and reflects the intrinsic efficiency of the fuzzing strategy. 
Lower values indicate more effective exploration and fewer redundant mutations.

\subsection{Implementation and Environment}
\label{subsec:implementation}

Regarding the selection of studied RL agents, we utilized an agent trained via the source code provided by RL Baselines3 Zoo ~\cite{rl-zoo3}for the MountainCar, while adopting the standard open-source agent commonly used in relevant baselines for the BipedalWalker~\cite{rl-zoo3}. For the CARLA autonomous driving environment, due to the unavailability of agent resources from prior studies, we deployed the Roach agent from the Pretrained CARLA Leaderboard Agents (PCLA)~\cite{PCLA} framework within the CARLA 0.9.15 simulator ~\cite{carla} as the subject under test.
To ensure experimental fairness and reproducibility, all baseline methods were configured strictly adhering to the optimal hyperparameters recommended by their respective authors~\cite{mdpfuzz,mdpfuzznew, QDfuzz,seqfuzz,curefuzz,g-model}. Detailed implementation specifics and source code have been made publicly available on our homepage. 
Experiments for MountainCar and BipedalWalker were conducted in a Python 3.10 environment using PyTorch 2.7.1~\cite{pytorch}. MountainCar utilized Stable-Baselines3 2.7~\cite{sb3} and Gymnasium 1.2.2~\cite{gym}, while BipedalWalker employed legacy libraries (Gym 0.19.0, Stable-Baselines3 1.1.0) consistent with prior work~\cite{mdpfuzz,mdpfuzznew, QDfuzz,seqfuzz,curefuzz,g-model}. In contrast, the CARLA environment was configured in strict adherence to PCLA specifications~\cite{PCLA}. To ensure statistical reliability, all experiments were executed five times on the same hardware using distinct random seeds.

All experiments were conducted on a server running Ubuntu 20.04, utilizing a computing infrastructure equipped with an Intel Xeon E5-2660 v4 CPU (128 GB RAM) and four NVIDIA GeForce RTX 2080 Ti GPUs.

\section{Experimental Results}
\label{sec:results_and_analysis}
\subsection{RQ1: Effectiveness}

\newcommand{\esN}{\textsuperscript{\scriptsize N}}
\newcommand{\esS}{\textsuperscript{\scriptsize S}}
\newcommand{\esM}{\textsuperscript{\scriptsize M}}
\newcommand{\esL}{\textsuperscript{\scriptsize L}}
\newcommand{\esRef}{\textsuperscript{\scriptsize ref}}

\begin{table}[!t]
\centering
\caption{Statistical Test Analysis with Effect Sizes}
\label{table:test}
\begin{adjustbox}{width=\columnwidth,center}
\begin{threeparttable}
\begin{tabular}{l|ccccc}
\toprule
\textbf{Method} &
\textbf{UC} &
\textbf{ID} &
\textbf{OD} &
\textbf{UD-AUC} &
\textbf{AG/UC} \\
\midrule
Random   & b\esL      & a\esM    & a\esN    & b\esL      & - \\
MDPFuzz  & a\esRef    & a\esN    & b\esL    & a\esRef    & a\esRef \\
CureFuzz & d\esL      & a\esL    & b\esL    & d\esL      & b\esL \\
G-Model  & c\esL      & a\esL    & a\esS    & c\esL      & c\esL \\
SeqDivFuzz  & e\esL      & a\esN    & b\esL    & d\esL      & b\esS \\
QDFuzz   & b\esL      & a\esRef  & a\esRef  & b\esL      & a\esN \\
\bottomrule
\end{tabular}

\begin{tablenotes}[flushleft]
\scriptsize
\item[*] Letters denote statistical groups. Superscripts denote effect-size magnitude relative to the reference: N = negligible, S = small, M = medium, L = large, ref = reference.

\end{tablenotes}

\end{threeparttable}
\end{adjustbox}
\vspace{-0.8em}
\end{table}

\begin{table*}[t]
\centering
\caption{Comparison of Fuzzing-Generated  Test Cases, UC, and Crash Ratios Across Benchmarks}
\label{table:test_crash_ratio_all}
\begin{adjustbox}{width=\textwidth,center}
\begin{threeparttable}
\begin{tabular}{c|ccc|ccc|ccc}
\toprule
\multirow{2}{*}{Method}
& \multicolumn{3}{c|}{MountainCar}
& \multicolumn{3}{c|}{BipedalWalker}
& \multicolumn{3}{c}{CARLA} \\
\cmidrule(lr){2-4} \cmidrule(lr){5-7} \cmidrule(lr){8-10}
& \#Total Test Cases & \#UC & Ratio
& \#Total Test Cases & \#UC & Ratio
& \#Total Test Cases & \#UC & Ratio \\
\midrule

Random
& $\bm{626,885 \pm 1.2\%}$ & $5,731 \pm 2.7\%$ & $0.91\% \pm 2.3\%$
& $\bm{444,764 \pm 0.6\%}$ & $8,026 \pm 0.9\%$ & $1.81\% \pm 0.5\%$
& $2,377 \pm 0.3\%$ & $157 \pm 4.5\%$ & $6.61\% \pm 4.6\%$ \\

MDPFuzz
& $446,376 \pm 1.5\%$ & $\bm{5,861 \pm 1.8\%}$ & $\bm{1.31\% \pm 1.1\%}$
& $359,301 \pm 0.2\%$ & $\bm{11,945 \pm 9.1\%}$ & $3.32\% \pm 9.0\%$
& $1,909 \pm 33.1\%$ & $\bm{291 \pm 55.1\%}$ & $\bm{15.19\% \pm 42.3\%}$ \\

CureFuzz
& $86,480 \pm 0.7\%$ & $1,045 \pm 2.0\%$ & $1.21\% \pm 2.2\%$
& $87,379 \pm 2.5\%$ & $1,000 \pm 0.0\%$ & $1.14\% \pm 2.6\%$
& $2,217 \pm 1.1\%$ & $140 \pm 41.6\%$ & $6.33\% \pm 41.5\%$ \\

G-Model
& $247,001 \pm 0.3\%$ & $1,210 \pm 3.3\%$ & $0.49\% \pm 3.2\%$
& $100,906 \pm 0.8\%$ & $1,050 \pm 4.6\%$ & $1.04\% \pm 4.5\%$
& $\bm{2,380 \pm 0.7\%}$ & $188 \pm 9.4\%$ & $7.92\% \pm 9.6\%$ \\

SeqDivFuzz
& $86,868 \pm 0.4\%$ & $1,048 \pm 3.7\%$ & $1.21\% \pm 4.0\%$
& $7,413 \pm 2.6\%$ & $97 \pm 7.2\%$ & $1.32\% \pm 7.7\%$
& $2,067 \pm 3.8\%$ & $77 \pm 66.9\%$ & $3.81\% \pm 71.8\%$ \\

QDFuzz
& $453,012 \pm 1.0\%$ & $5,455 \pm 1.6\%$ & $1.20\% \pm 0.9\%$
& $131,919 \pm 0.6\%$ & $7,330 \pm 3.4\%$ & $\bm{5.56\% \pm 3.6\%}$
& $2,175 \pm 1.0\%$ & $195 \pm 68.2\%$ & $9.02\% \pm 69.0\%$ \\

\bottomrule
\end{tabular}
\begin{tablenotes}
    \footnotesize
    \item[*] Data is presented as $a\pm b\%$, where $a$ is the mean value and $b\%$ is the relative standard deviation (RSD). The highest result within each benchmark and metric is highlighted in bold.
\end{tablenotes}
\end{threeparttable}
\end{adjustbox}
\end{table*}

To investigate performance discrepancies, we employ the Wilcoxon Signed-Rank test~\cite{wilcoxon,DBLP:journals/stvr/ArcuriB14} (with a significance level of 0.05) combined with the Benjamini-Hochberg correction method~\cite{benjamini1995controlling} to detect statistically significant differences, in line with the evaluation pipeline used in previous research~\cite{DBLP:journals/tosem/ZhouCH22,zheng2025identifying,fixstudy}. Methods are divided into the same groups when no significant difference exists between them, and marked with alphabetical labels $\{\textit{a}, \textit{b}, \textit{c}\}$ where earlier labels represent better performance (i.e., \textit{a} > \textit{b} > \textit{c}). The corresponding results are summarized in Table~\ref{table:test}. We further compute the rank-biserial correlation coefficient~\cite{cureton1956rank}$r$ to measure the magnitude of observed differences. Following existing work~\cite{cohen1988statistical}, the effect size is interpreted by magnitude: $r \ge 0.5$ indicates a large effect, $0.3 \le r < 0.5$ indicates a medium effect, $0.1 \le r < 0.3$ indicates a small effect, and $r < 0.1$ indicates a negligible effect.

Table~\ref{table:test_crash_ratio_all} presents the number of UCs detected by the six fuzzing methods on three RL benchmarks. 
Overall, MDPFuzz achieves the best performance across all tasks. Notably, it is the only method assigned to group \textit{a} (Table~\ref{table:test}), which denotes a statistically significant performance advantage over all other methods.
Taking the BipedalWalker benchmark as an example, MDPFuzz detects 11,945 crashes, representing a 48.82\% improvement over random testing, a 62.96\% improvement over QDFuzz, and more than a 100-fold increase compared to SeqDivFuzz. 
This advantage is largely attributed to its lightweight design and high computational efficiency, enabling rapid mutation and exploration within the fixed time budget. 
In contrast, SeqDivFuzz exhibits the weakest overall performance, detecting the fewest crashes on two benchmarks (97 on BipedalWalker and 77 on CARLA, marked as group \textit{e} in Table~\ref{table:test}). 
A likely explanation is that SeqDivFuzz relies primarily on state coverage as its exploration guidance.
While this strategy promotes broader state exploration, prior studies~\cite{mdpfuzznew,DBLP:conf/icse/TrujilloLEDC20} suggest that coverage alone provides limited correlation with crash-triggering behaviors, thereby reducing its ability to efficiently expose crashes.

\begin{tcolorbox}\textbf{Finding I:}
MDPFuzz consistently outperforms all compared methods across benchmarks, demonstrating superior crash detection capability largely due to its high computational efficiency and fast exploration. 
On the other hand, SeqDivFuzz shows comparatively weaker overall performance, as its state-coverage–based exploration guidance provides limited effectiveness in prioritizing crash-triggering behaviors.
\end{tcolorbox}

On low-dimensional MountainCar and moderately complex BipedalWalker, random testing ranks second only to MDPFuzz.
This can be explained by their low-dimensional state spaces: faulty states are generally easy to find without complex guidance.
Therefore, methods that rapidly generate test inputs (e.g., random testing), benefit from higher testing throughput, increasing the likelihood of discovering crashes within the fixed time budget.
While methods with computationally intensive components may underperform due to reduced mutation rates. For example, CureFuzz requires training a Random Network Distillation module~\cite{RND}, and G-Model depends on diffusion-based modelling of test input distributions~\cite{slow}, both of which introduce additional overhead.
As task complexity increases, guided exploration becomes critical.
For example, QDFuzz ranks third on the MountainCar and BipedalWalker benchmarks but rises to second place on CARLA, outperforming random testing. This demonstrates that random testing becomes less effective in large search spaces, while quality-diversity search enables QDFuzz to cover diverse crash modes and achieve stronger performance in complex scenarios.
Nevertheless, random testing maintains remarkable performance in producing crash-triggering test inputs, holding group \textit{b} in Table~\ref{table:test} and outperforming many methods, highlighting that random testing should be retained as a key baseline.



\begin{tcolorbox}\textbf{Finding II:}
For low-dimensional and moderately complex RL tasks where target agents typically exhibit lower robustness, highly efficient fuzzing methods (e.g., random testing) tend to yield better performance. 
As task complexity increases, effective exploration guidance becomes critical for discovering crashes, and QDFuzz exhibits notably improved performance in such complex scenarios by leveraging effective exploration guidance strategies.

\end{tcolorbox}

\subsection{RQ2: Diversity}

\begin{table}[t]
\centering
\caption{Comparison of ID and OD Across Benchmarks}
\label{table:input_output_diversity}
\begin{adjustbox}{width=0.5\textwidth,center}
\begin{threeparttable}
\begin{tabular}{c|cccccc}
\toprule
\multirow{2}{*}{Method}
& \multicolumn{2}{c}{MountainCar}
& \multicolumn{2}{c}{BipedalWalker}
& \multicolumn{2}{c}{CARLA} \\
\cmidrule(lr){2-3} \cmidrule(lr){4-5} \cmidrule(lr){6-7}
& Input & Output
& Input & Output
& Input & Output \\
\midrule

Random     & $2.0 \pm 0.0\%$  & $2.0 \pm 0.0\%$  & $2.0 \pm 0.0\%$  & $15.8 \pm 15.1\%$ & $2.0 \pm 0.0\%$   & $6.2 \pm 7.2\%$  \\
MDPFuzz    & $2.0 \pm 0.0\%$  & $2.0 \pm 0.0\%$  & $2.0 \pm 0.0\%$  & $13.0 \pm 9.4\%$  & $4.0 \pm 46.8\%$  & $2.0 \pm 0.0\%$ \\
CureFuzz   & $2.4 \pm 37.3\%$ & $2.0 \pm 0.0\%$  & $2.0 \pm 0.0\%$  & $2.0 \pm 0.0\%$   & $\bm{12.0 \pm 15.6\%}$ & $7.4 \pm 71.3\%$ \\
G-Model    & $2.0 \pm 0.0\%$  & $2.0 \pm 0.0\%$  & $2.0 \pm 0.0\%$  & $2.0 \pm 0.0\%$   & $2.0 \pm 0.0\%$   & $\bm{10.4 \pm 27.7\%}$ \\
SeqDivFuzz & $\bm{2.8 \pm 39.1\%}$ & $2.0 \pm 0.0\%$ & $\bm{8.0 \pm 75.5\%}$ & $2.0 \pm 0.0\%$ & $10.0 \pm 27.4\%$ & $8.4 \pm 57.5\%$ \\
QDFuzz     & $2.4 \pm 37.3\%$ & $2.0 \pm 0.0\%$  & $2.0 \pm 0.0\%$  & $\bm{16.4 \pm 12.6\%}$ & $4.0 \pm 25.0\%$ & $4.2 \pm 66.1\%$ \\
\bottomrule

\end{tabular}
\end{threeparttable}
\end{adjustbox}
\end{table}

Table~\ref{table:input_output_diversity} reports the input and output diversity of each fuzzing method across the three benchmarks, presented as mean and RSD values.
The top results are highlighted in bold.
The overall results show that no single method can achieve an absolute leading performance in all benchmarks.
For example, the methods ranking first in ID on the three benchmarks are SeqDivFuzz (2.8), SeqDivFuzz (8), and CureFuzz (12), respectively; the top performers in OD on BipdedalWalker and CARLA are QDFuzz (16.4) and G-Model (10.4) in turn. 
This indicates that the diversity performance of fuzzing methods is highly dependent on the tested benchmarks.

Among all fuzzing methods, SeqDivFuzz demonstrates relatively strong diversity performance across multiple benchmarks. It ranks first on two diversity metrics, achieving the highest ID on MountainCar (2.8) and the highest OD on BipedalWalker (8).
This is likely due to its adopted \textit{golden sequence sampling} and \textit{k-voting mechanism}, which determine whether to terminate or continue the current test case based on the similarity between the current and golden state sequences. This mechanism switches between different seeds for fuzzing and generates diverse test inputs. 
We further observe that MDPFuzz, despite uncovering the largest number of crash-triggering test inputs, exhibits poor diversity. Its OD score on CARLA stands at merely 2.0, even lower than random testing. This indicates the crashes it triggers are highly similar.



\begin{figure}[t]
    \centering
    \includegraphics[width=1.0\linewidth]{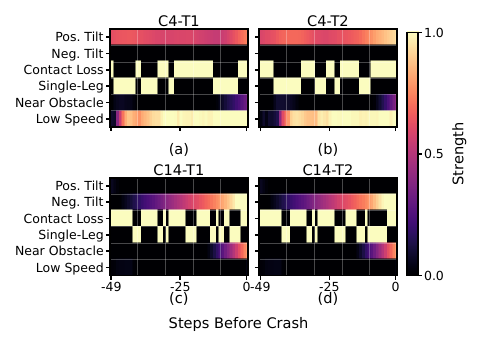}
    \caption{Examples from BipedalWalker for Qualitative Analysis}
    \label{fig:heatmap}
\end{figure}

To explore whether the proposed metric can effectively trigger diverse faulty behaviors, we conduct a qualitative analysis following existing work~\cite{DBLP:conf/icse/YouWCLL23}. For clear visualization and easy presentation, we sample a subset of test inputs for analysis. We further adopt a concept-based explanation method~\cite{DBLP:conf/icml/KohNTMPKL20} to generate heatmaps to measure how each input dimension contributes to behavioral decisions of each step. Figure~\ref{fig:heatmap} presents the heatmaps of crash-triggering trajectories, where (a) and (b) are sampled from the same cluster, and (c) and (d) belong to another cluster. The results show that trajectories within the same cluster share highly similar root causes of faulty behaviors. Specifically, the robot crashes illustrated in Figure~\ref{fig:heatmap}(a) and ~\ref{fig:heatmap}(b) are due to the unstable support state during low-speed movement, which eventually leads to forward falls. In contrast, the faulty behaviors in Figure~\ref{fig:heatmap}(c) and ~\ref{fig:heatmap}(d) result from posterior posture instability and failed support recovery, causing backward falls, which are different from the crash patterns of (a) and (b).
The above phenomena indicate that trajectories from different clusters correspond to different root causes. This further shows that the cluster-based diversity criterion can reflect the diversity of crash-triggering trajectories to some degree.

\begin{tcolorbox}\textbf{Finding III:}
 Fuzzing diversity varies widely across benchmarks; no method dominates both ID and OD. SeqDivFuzz shows relatively better diversity performance on specific benchmarks due to its specialized mechanisms, while MDPFuzz yields limited diversity with mostly similar crashes.
\end{tcolorbox}



\begin{figure}[t]
    \centering
    \includegraphics[width=1.0\linewidth]{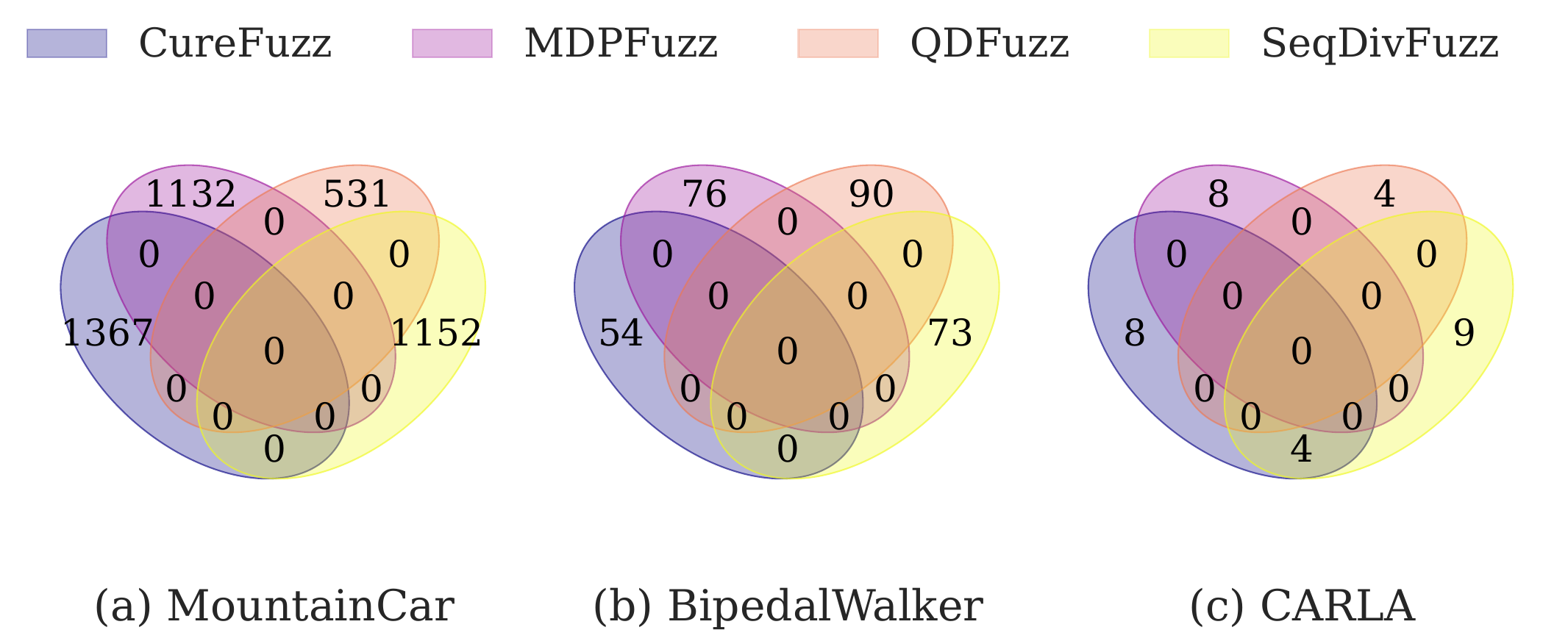}
    \caption{Overlap of Initial Seeds for UC Detected}
    \label{fig:venn}
\end{figure}

Figure~\ref{fig:venn} illustrates that the sets of crash-inducing initial seeds identified by different fuzzing methods exhibit extremely low overlap, indicating that different methods rarely discover crashes from the same starting conditions. 
This observation suggests that the fuzzers follow different exploration trajectories within the state space and tend to uncover distinct regions of crash behaviors. Consequently, no single method is sufficient to cover all testing blind spots, highlighting the complementary nature of diverse fuzzing strategies and the potential benefits of combining multiple methods for more comprehensive crash detection.

\begin{tcolorbox}\textbf{Finding IV:}
Current fuzzing methods exhibit extremely low overlap in crash-inducing initial seeds, indicating that they explore different regions of the state space and uncover distinct crash behaviors.
\end{tcolorbox}







\subsection{RQ3: Efficiency}


\begin{figure}[tbp]
    \centering
    \begin{subfigure}{0.32\linewidth}
        \centering
        \includegraphics[width=\linewidth, height=2.5cm]{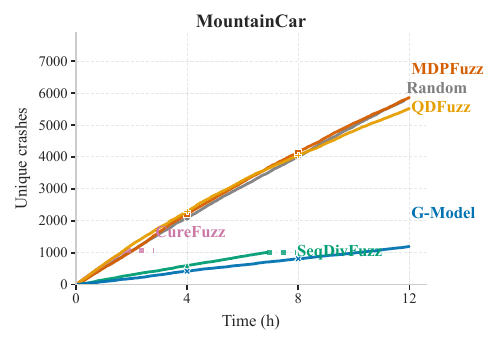} 
        \caption{MountainCar} 
        \label{subfig:rq3.1-agpuc-1}
    \end{subfigure}
    \hfill 
    \begin{subfigure}{0.32\linewidth}
        \centering
        \includegraphics[width=\linewidth,height=2.5cm]{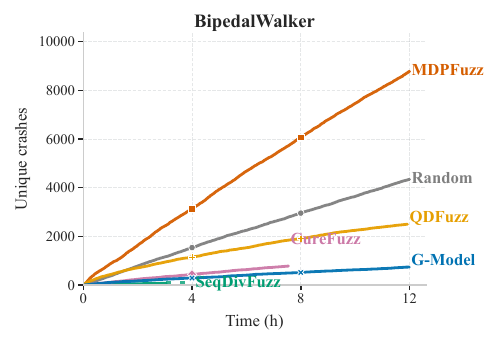}  
        \caption{BipedalWalker}
        \label{subfig:rq3.1-agpuc-2}
    \end{subfigure}
    \hfill
    \begin{subfigure}{0.32\linewidth}
        \centering
        \includegraphics[width=\linewidth,height=2.5cm]{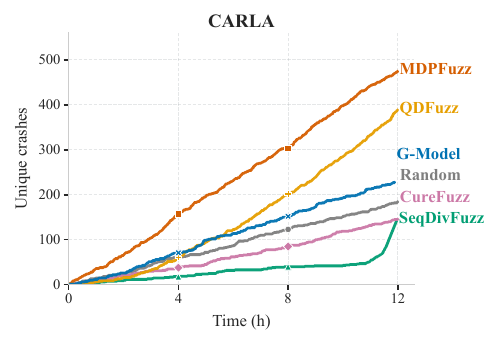}  
        \caption{CARLA}
        \label{subfig:rq3.1-agpuc-3}
    \end{subfigure}
    \caption{Temporal Trends of UC Numbers Detected by Different Fuzzing Methods}
    \label{fig:rq3.1-agpuc}
\end{figure}

\begin{table}[t]
\centering
\caption{Comparison of UD-AUC Across Benchmarks}
\label{table:unique_crash_auc_comparison}
\begin{adjustbox}{width=0.5\textwidth,center}
\begin{threeparttable}
\begin{tabular}{c|ccc}
\toprule
Method & MountainCar & BipedalWalker & CARLA \\
\midrule
Random     & $35329.8 \pm 3.1\%$           & $48439.4 \pm 1.4\%$           & $937.2 \pm 5.4\%$             \\
MDPFuzz    & $\bm{38280.3 \pm 2.4\%}$  & $\bm{77814.9 \pm 8.3\%}$  & $\bm{1621.7 \pm 58.7\%}$  \\
CureFuzz   & $942.9 \pm 3.2\%$             & $1473.8 \pm 2.6\%$            & $813.8 \pm 32.8\%$            \\
G-Model    & $7246.0 \pm 3.7\%$            & $6626.5 \pm 4.8\%$            & $1137.4 \pm 8.7\%$            \\
SeqDivFuzz & $3609.7 \pm 5.7\%$            & $118.5 \pm 12.1\%$            & $392.9 \pm 34.9\%$            \\
QDFuzz     & $36555.8 \pm 1.9\%$           & $54275.1 \pm 3.6\%$           & $993.1 \pm 65.2\%$            \\
\bottomrule
\end{tabular}
\end{threeparttable}
\end{adjustbox}
\end{table}

\smallskip
\noindent
\textbf{Results.} 
Figure~\ref{fig:rq3.1-agpuc} shows the cumulative number of unique crashes discovered over time by different fuzzing methods across the three benchmarks, while Table~\ref{table:unique_crash_auc_comparison} reports UD-AUC. Overall, MDPFuzz consistently demonstrates the highest time efficiency on all benchmarks, substantially outperforming the competing methods.
This trend is corroborated by Figure~\ref{fig:rq3.1-agpuc}, where the MDPFuzz curve remains consistently above others and widens its lead as task complexity increases, indicating faster and more sustained crash discovery.

Taking the BipedalWalker benchmark as an example, MDPFuzz outperforms QDFuzz by 43.4\% and surpasses SeqDivFuzz by over 600 times. This advantage stems from two main factors.
First, its lightweight exploration guidance incurs minimal computational overhead while effectively prioritizing crash-prone regions. In contrast, methods such as SeqDivFuzz spend substantial time computing coverage information, which reduces mutation throughput and leads to higher time costs.
Second, MDPFuzz permits the seed pool to grow dynamically, enabling continuous and sustained exploration. By comparison, CureFuzz and SeqDivFuzz employ aggressive seed pruning strategies that restrict seed lifecycles, causing their fuzzing processes to terminate prematurely. 
As shown in Figure~\ref{fig:rq3.1-agpuc}(a) and Figure~\ref{fig:rq3.1-agpuc}(b), both methods exhaust their exploration within four hours on MountainCar and eight hours on BipedalWalker, respectively.


\begin{tcolorbox}\textbf{Finding V:}
MDPFuzz achieves the highest crash detection efficiency. 
In contrast, SeqDivFuzz suffers from lower efficiency because of its time-consuming coverage calculations, while both CureFuzz and SeqDivFuzz often terminate fuzzing prematurely due to restrictive seed lifecycle management.
\end{tcolorbox}



\begin{figure}[tbp]
    \centering
    \begin{subfigure}{0.32\linewidth}
        \centering
        \includegraphics[width=\linewidth]{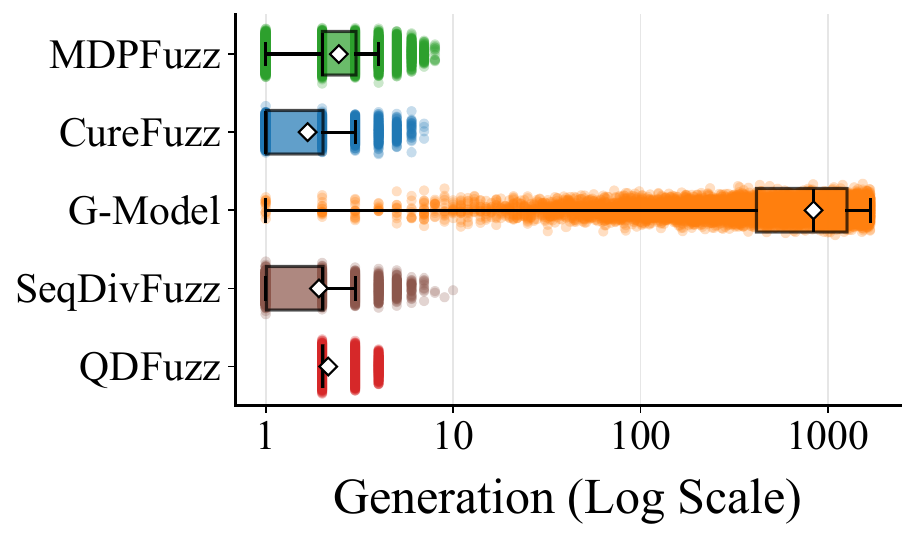} 
        \caption{MountainCar} 
        \label{subfig:rq3-agpuc-1}
    \end{subfigure}
    \hfill 
    \begin{subfigure}{0.32\linewidth}
        \centering
        \includegraphics[width=\linewidth]{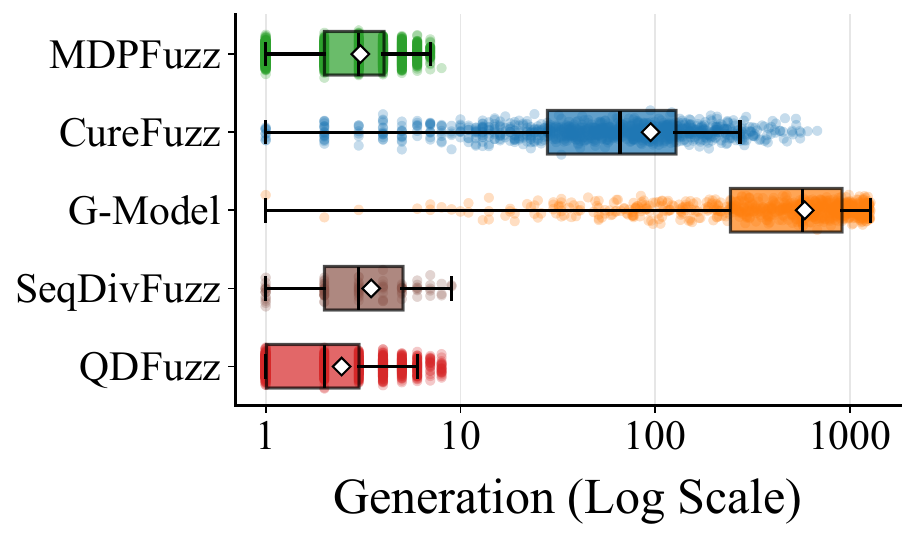}  
        \caption{BipedalWalker}
        \label{subfig:rq3-agpuc-2}
    \end{subfigure}
    \hfill
    \begin{subfigure}{0.32\linewidth}
        \centering
        \includegraphics[width=\linewidth]{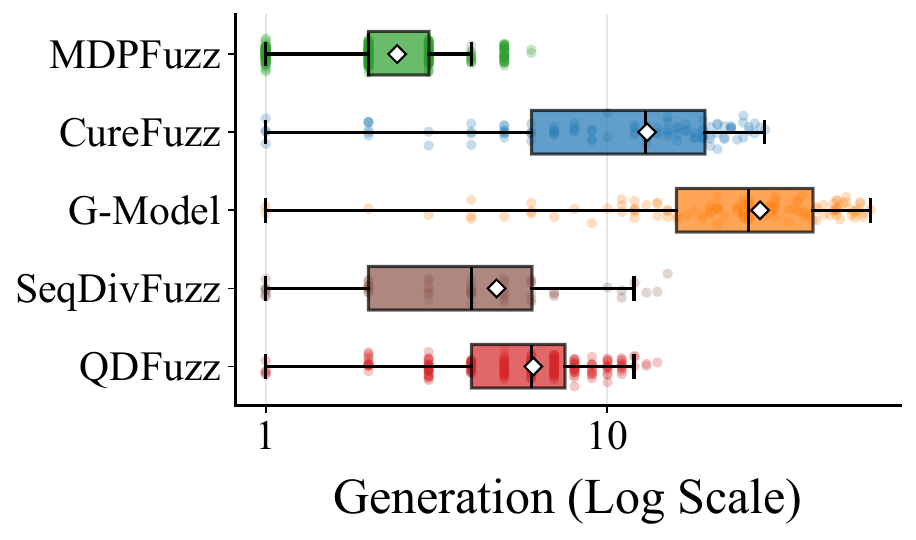}  
        \caption{CARLA}
        \label{subfig:rq3-agpuc-3}
    \end{subfigure}
    \caption{Box-plot Analysis of AG/UC for Different Fuzzers}
    \label{fig:rq3-agpuc}
\end{figure}

Figure~\ref{fig:rq3-agpuc} presents the distribution of mutation generations required by different fuzzing methods to generate crash-triggering test inputs (i.e., generations per unique crash) across the three benchmarks.
The horizontal axis shows the logarithmically transformed number of iterations to accommodate the wide variation in scales. 
In this visualization, a leftward shift of a boxplot indicates fewer required iterations and thus higher crash detection efficiency, while the white squares denote average values.
Overall, MDPFuzz, QDFuzz, and SeqDivFuzz consistently yield the lowest values of generations per unique crash, demonstrating superior search efficiency. 
For example, on MountainCar (Figure~\ref{subfig:rq3-agpuc-1}), these methods require approximately one mutation on average to trigger a crash, whereas the generative method G-Model requires nearly 1,000 iterations. 
This substantial gap suggests that mutation-based methods can rapidly expose crashes through lightweight heuristic guidance. 
In contrast, generative methods such as G-Model incur additional overhead to model or learn the behavioral distribution of test cases, which increases the number of iterations needed before crashes can be effectively triggered.

\begin{tcolorbox}\textbf{Finding VI:}
MDPFuzz, QDFuzz, and SeqDivFuzz outperform other methods across three benchmarks with the lowest generations per unique crash via heuristic guidance, while the generation-based method G-Model requires far more iterations for pre-learning the behavioral distribution of tested RL agents.
\end{tcolorbox}

\subsection{RQ4: Usefulness}

We further investigate the practical utility of the crash-triggering test inputs generated by each fuzzing method by designing experiments around two representative downstream tasks commonly studied in prior work: agent robustness improvement and safety monitoring. Both tasks involve additional agent training and thus incur non-trivial computational costs.
We carefully select the evaluation benchmark to balance feasibility and representativeness. 
The MountainCar environment is relatively simple, which can lead to overfitting during training and limit the meaningfulness of robustness evaluation. 
In contrast, CARLA is substantially more complex and computationally demanding, requiring over one month of training time in our experimental setting. 
Therefore, we adopt BipedalWalker, a moderately complex environment, as a proof-of-concept benchmark that provides sufficient task difficulty while remaining computationally tractable.

\textbf{I. Agent robustness improvement } aims to leverage crash-triggering test inputs to enhance the resilience of RL agents against crashes. We adopt the experiment pipeline of prior work~\cite{mdpfuzz}. Specifically, we first collect all crash-triggering cases discovered during 12 hours of fuzzing from the same initial seed corpus. 
The corresponding five-element tuples are then incorporated into the replay buffer with higher sampling weights to fine-tune the policy, encouraging the agent to learn from previously crash-prone scenarios. We adopt this strategy for two primary reasons: 1) it has been evaluated and adopted in existing literature~\cite{DBLP:conf/smc/HouLWXC17,DBLP:SchaulQAS15}, ensuring experimental replicability and result comparability;
2) it features a lightweight mechanism~\cite{DBLP:journals/corr/abs-2601-02648} that can be readily adapted to our framework. In contrast, the methods proposed in other work~\cite{DBLP:conf/icse/TapplerPAK24} rely on complex designs and scenario-specific pipelines, which is difficult to generalize and transfer to our experimental setting.
After fine-tuning, we launch a second 12-hour fuzzing run on the repaired agent with unchanged initial seeds and hyperparameters. We quantify robustness gains by contrasting unique crash counts pre- and post-fine-tuning. Fewer detected crashes indicate stronger robustness.

In addition, to eliminate self-evaluation bias, we further perform cross-fuzzer evaluation. Specifically, we repair the policy using crashes from one fuzzer and assess robustness via separate fuzzers. For both self- and cross-fuzzer evaluation, we use the same replay buffer size to ensure fair comparisons. Each holds 10k crash transitions and 90k normal transitions. Raw crash cases are decomposed into individual transitions for storage and not tallied as standalone samples. Normal transitions are uniformly sampled from the agent’s original training data, while crash transitions stem from crashes discovered by each fuzzer under identical initial seeds. Crash transitions are assigned a sampling weight of 10 versus a weight of 1 for normal transitions, balancing general behavioral learning and crash awareness. All fine-tuning procedures adopt unified hyperparameters from existing research~\cite{DBLP:SchaulQAS15,DBLP:abs-2005-01643,DBLP:abs-2303-17396} learning rate = \(10^{-7}\), with 50k offline gradient updates.

\textbf{II. Safety monitoring} is a task specific to RL, aiming to predict whether an ongoing episode is likely to result in a crash based on historical state–action information. We follow the experimental configuration of prior work~\cite{DRLFailureMonitor}.
Specifically, for each test case, we record the observations and actions at every interaction step between the agent and the environment and concatenate them into sequential feature vectors. 
Based on the final execution outcome, each test case is labelled as either crash-triggering or normal. 
To construct representative inputs, we apply differentiated sequence extraction: for crash-triggering cases, we select the last 25 steps preceding the crash, whereas for successful executions, we randomly sample a sequence of the same length to ensure consistency.

After collecting the data, we perform balancing and random sampling following recommended practices, constructing a dataset of 3,000 samples with 30\% crash cases.
The dataset is then preprocessed through transposition, shuffling, and dimensional expansion to meet the input requirements of the agent and is split into training and validation sets with a 7:3 ratio. Finally, we train a TodyNet~\cite{liu2024todynet} classifier to predict whether a test case is likely to trigger a crash.
To assess generalization, we evaluate not only the within-method prediction accuracy, i.e., training and testing on crashes generated by the same fuzzing method, but also cross-method performance, where the trained agent is tested on crashes discovered by other fuzzers. 
This setting enables us to examine whether the detected crash patterns generalize across different fuzzing strategies.


\begin{table}[t]
\centering
\caption{Self-fuzzer Evaluation of Robustness Improvement}
\label{table:retraining_impact}
\begin{adjustbox}{width=0.5\textwidth,center}
\begin{threeparttable}
\begin{tabular}{c|ccccc}
\toprule
Method & \#UC Before Fine-tuning & \#UC After Fine-tuning & $\Uparrow_{Rob}$ & F.R.$(\Downarrow_{F.R.})$ & $Reg.$ \\
\midrule
Random     & 4,399 & 4,044 & 8.1\%            & 2.6\%(-1.0) & 2.0\% \\
MDPFuzz    & 8,952 & 7,203 & 19.5\%           & 1.3\%(+0.3) & 0.9\% \\
CureFuzz   & 804   & 804   & 0.00\%            & 1.7\%(-0.1) & 1.3\% \\
G-Model    & 773   & 664   & 14.1\%            & 1.3\%(+0.3) & 1.1\% \\
SeqDivFuzz & 85    & 75    & 11.8\%            & 1.6\%(+0) & 1.4\% \\
QDFuzz     & 2,418 & 1,414 & 41.5\%  & 1.9\%(-0.3) & 1.5\% \\
\bottomrule
\end{tabular}
\begin{tablenotes}
\footnotesize
\item[*] \textit{Note:} $\Uparrow_{Rob}$ denotes the robustness improvement rate; F.R. denotes the failure rate, with the value in parentheses indicating its change after fine-tuning; Reg. denotes the regression error rate.
\end{tablenotes}
\end{threeparttable}
\end{adjustbox}
\end{table}

\begin{table}[t]
\centering
\caption{Cross-Fuzzer Evaluation of Robustness Improvement (\#UC after Fine-tuning and $\Uparrow_{Rob}$)}
\label{table:cross_fuzzing_evaluation}
\begin{adjustbox}{width=0.5\textwidth,center}

\begin{threeparttable}

\begin{tabular}{c|llllll}
\toprule
Tra.\textbackslash{}Eva. & Random & MDPFuzz & CureFuzz & G-Model & SeqDivFuzz & QDFuzz \\
\midrule
Random     & 4,044 (8.1\%)   & 6,763 (24.5\%)  & 802 (0.2\%)  & 712 (7.9\%)  & 98 (-15.3\%)  & 1,179 (51.2\%) \\
MDPFuzz    & 3,927 (10.7\%)  & 7,203 (19.5\%)  & 798 (0.7\%)  & 703 (9.1\%)  & 105 (-23.5\%) & 983 (59.3\%)   \\
CureFuzz   & 6,502 (-47.8\%) & 11,291 (-26.1\%) & 804 (0.0\%) & 672 (13.1\%) & 100 (-17.6\%) & 1,023 (57.7\%) \\
G-Model    & 5,613 (-27.6\%) & 10,323 (-15.3\%) & 785 (2.4\%) & 664 (14.1\%) & 100 (-17.6\%) & 2,487 (-2.9\%) \\
SeqDivFuzz & 3,629 (17.5\%)  & 8,353 (6.7\%)   & 788 (2.0\%)  & 681 (11.9\%) & 75 (11.8\%)   & 496 (79.5\%)   \\
QDFuzz     & 5,062 (-15.1\%) & 11,702 (-30.7\%) & 792 (1.5\%) & 622 (19.5\%) & 104 (-22.4\%) & 1,414 (41.5\%) \\
\bottomrule
\end{tabular}

\begin{tablenotes}
\footnotesize
\item[*] Rows denote the fuzzers whose UCs are used to fine-tune the RL agent, while columns denote the fuzzers used for evaluation. Each cell reports the number of UCs detected after fine-tuning, with the robustness improvement rate ($\Uparrow_{Rob}$) shown in parentheses. The improvement rate measures the percentage reduction of UCs compared with the initial number of UCs, where the pre-fine-tuning results are reported in the column \textit{\#UC Before Fine-tuning} of Table~\ref{table:retraining_impact}. A negative value indicates robustness degradation.
\end{tablenotes}

\end{threeparttable}
\end{adjustbox}
\end{table}
\begin{table}[t]
\centering
\caption{Safety Monitoring Accuracy Based on Crash-Triggering Inputs Across Fuzzing Methods}
\label{table:safety}
\begin{adjustbox}{width=0.5\textwidth,center}

\begin{threeparttable}{

\begin{tabular}{c|clllll}

\toprule
Tra.\textbackslash{}Eva. & Random   & MDPFuzz  & CureFuzz & G-Model  & SeqDivFuzz & QDFuzz   \\ \midrule
Random                             & 100.00\% & 99.67\%  & 93.17\%  & 93.50\% & 94.50\%    & 99.17\%  \\
MDPFuzz                            & 100.00\% & 100.00\% & 93.67\%  & 93.83\% & 95.17\%    & 99.17\%  \\
CureFuzz                           & 99.67\%  & 99.50\%  & 99.67\%  & 95.00\% & 98.50\%    & 97.50\%  \\
G-Model                             & 97.33\%  & 99.83\%  & 99.00\%  & 96.83\% & 99.00\%    & 98.83\%  \\
SeqDivFuzz                         & 99.83\%  & 99.67\%  & 99.50\%  & 96.33\% & 97.83\%    & 98.50\%  \\
QDFuzz                            & 97.33\%  & 99.67\%  & 93.50\%  & 94.00\% & 95.00\%    & 100.00\% \\ 
\bottomrule
\end{tabular} 
}

\end{threeparttable}
\end{adjustbox}
\end{table}

\begin{table}[t]
\centering
\caption{Safety Monitoring False Positive Rate Based on Crash-Triggering Inputs Across Fuzzing Methods}
\label{table:fpr_percent}
\begin{adjustbox}{width=0.5\textwidth,center}

\begin{threeparttable}
\begin{tabular}{c|cccccc}
\toprule
Tra.\textbackslash{}Eva. & Random  & MDPFuzz & CureFuzz & G-Model & SeqDivFuzz & QDFuzz  \\ \midrule
Random      & 0.00\%  & 0.24\%  & 0.24\%   & 0.23\%  & 0.24\%     & 0.00\%  \\
MDPFuzz     & 0.23\%  & 0.00\%  & 0.00\%   & 0.23\%  & 0.00\%     & 0.00\%  \\
CureFuzz    & 0.46\%  & 0.72\%  & 0.24\%   & 5.16\%  & 1.67\%     & 3.48\%  \\
G-Model     & 0.46\%  & 0.24\%  & 0.24\%   & 2.35\%  & 0.48\%     & 1.62\%  \\
SeqDivFuzz  & 0.23\%  & 0.48\%  & 0.24\%   & 3.76\%  & 1.91\%     & 2.09\%  \\
QDFuzz      & 0.23\%  & 0.24\%  & 0.00\%   & 0.23\%  & 0.00\%     & 0.00\%  \\ \bottomrule
\end{tabular}
\end{threeparttable}

\end{adjustbox}
\end{table}

Related to \textbf{Agent robustness improvement,} 
Table~\ref{table:retraining_impact} reports the number of unique crashes detected during 12-hour fuzzing sessions under identical datasets and configurations before and after agent repair.
The $\Uparrow_{Rob}$ quantifies the reduction in detected crashes and serves as an indicator of robustness improvement, and $F.R.$ indicates the overall accuracy of the fine-tuned RL agent, and $Reg.$ denotes the ratio of regression faults in the test sets~\cite{DBLP:journals/tosem/YouWCCSLD25}.
As shown in the results, QDFuzz achieves the most substantial gain, improving agent robustness by 41.5\%. This advantage is likely attributable to its emphasis on generating diverse and high-quality test cases, which expose a broader range of crash scenarios and thus provide more informative training signals during repair. More generally, test cases produced by most fuzzing methods contribute to varying degrees of robustness enhancement, demonstrating the practical value of fuzzing-generated crashes for agent improvement.
We also observe that fine-tuning on crashes from G-Model, CureFuzz, and random testing may degrade overall agent performance and introduce potential regression faults.


To evaluate whether fine-tuning with crashes generated by one fuzzer can improve robustness against attacks from other fuzzers, we conducted a cross-fuzzer evaluation. As shown in Table~\ref{table:cross_fuzzing_evaluation}, the rows denote the fuzzers used to generate the fine-tuning data, while the columns denote the fuzzers used for evaluation. Each cell reports the number of unique crashes after fine-tuning and the corresponding robustness improvement relative to the pre-fine-tuning results in Table~\ref{table:retraining_impact}. Positive values indicate improved robustness.
Fine-tuning with crashes generated by SeqDivFuzz consistently improves robustness against all evaluation fuzzers, with improvements ranging from 17.5\% to 79.5\%, demonstrating the strongest cross-fuzzer generalization ability. This advantage is likely due to its diversity-aware pruning strategy, which removes redundant crashes and produces more transferable repair signals. In contrast, fine-tuning with crashes generated by other fuzzers often degrades robustness against unseen fuzzers. For example, using crashes generated by CureFuzz reduces robustness against Random Testing by 47.8\%, indicating overfitting to the source fuzzer.

\begin{tcolorbox}\textbf{Finding VII:}
Most fuzzing methods can enhance the robustness of RL agents through repair based on their generated crash-triggering test cases, with QDFuzz yielding the largest gains and SeqDivFuzz showing the strongest cross-fuzzer generalization, yet such fine-tuning may introduce regression faults.
\end{tcolorbox}

Related to \textbf{Safety Monitoring,} 
Table~\ref{table:safety} and Table~\ref{table:fpr_percent} evaluate the accuracy and the false positive rate of safety monitoring models trained on crash-triggering inputs generated by different fuzzing methods in identifying crash-prone test cases. Specifically, we assess both within-method performance, where training and evaluation use crashes produced by the same fuzzer, and cross-method performance across different fuzzers.
The results show that within-method training typically achieves the highest recognition accuracy. For example, the model trained on MDPFuzz-generated crashes attains 100.00\% accuracy when evaluated on crashes detected by MDPFuzz itself. 
This outcome suggests that the model effectively learns the characteristic patterns associated with crashes produced by that specific fuzzing strategy.
Notably, even under cross-method evaluation, the monitoring models maintain high performance, with accuracies exceeding 95\% in most cases.  For example, the model trained on G-Model-generated crashes attains 97.33\%$\sim$99.83\% when evaluated on other fuzzing methods. We also observe that for most cases, the false positive rate is below 1.00\%, with a maximum of no more than 5.16\%. This strong generalization indicates that, despite method-specific differences, crash-triggering inputs share common behavioral signatures.
Such shared characteristics enable the trained models to reliably identify crash-prone scenarios across different fuzzing strategies, demonstrating the practical value of fuzzing-generated data for safety monitoring.

\begin{tcolorbox}\textbf{Finding VIII:}
Safety monitoring models trained on crash-triggering inputs from a single fuzzing method yield optimal accuracy on the same method’s crash inputs and maintain high accuracy in most cross-method evaluations, potentially owing to the core common crash behavior characteristics shared by crash inputs from different fuzzing methods, which grant the models strong cross-method generalization ability.

\end{tcolorbox}

\section{Lesson Learned and Actionable Takeaways}
Based on our comprehensive empirical study, we distil several key implications for both researchers and practitioners.

\textbf{Select Appropriate Fuzzers According to Testing Goals.} For simple tasks with small state spaces, low computational overhead is a primary concern; random testing and MDPFuzz are viable options. For high-dimensional, complex simulation tasks, random testing alone fails to uncover rare risky scenarios, so guided fuzzing methods, including MDPFuzz and QDFuzz are preferred.
For time-limited scenarios such as pre-release stress testing where maximum crash discovery within a fixed budget is required, MDPFuzz remains the optimal solution. It generates the largest number of UCs and achieves the best UD-AUC score, demonstrating its strong crash detection capacity and ability to reveal latent crashes early.
For generating rare, long-tail, and atypical abnormal behaviors, especially in offline deep testing on complex simulators like autonomous driving platforms, G-Model is recommended since it learns the distribution of normal scenarios and creates outliers, making it well-suited for mining long-tail risks.
QDFuzz and SeqDivFuzz are suitable when diverse crash samples are needed for robustness enhancement. They produce behaviorally distinct UCs to support agent repair and strengthen the repaired agent’s robustness to abnormal attacks.

\textbf{Combine Complementary Fuzzing Strategies.}
Our observations suggest that Different fuzzing methods tend to explore distinct regions of the state space, resulting in low overlap in crash-inducing seeds.
As shown in Figure~\ref{fig:venn}, the intersection of seeds across methods is nearly zero on all benchmarks, indicating that each strategy uncovers unique crash modes. This yields three key implications.
1) Relying solely on one fuzzer creates wide blind spots in state exploration. Practitioners may integrate complementary fuzzing pipelines to expand total crash coverage. A representative combination pairs efficiency-focused tools such as MDPFuzz with diversity-driven alternatives like QDFuzz, jointly boosting crash detection throughput and behavioral coverage.
2) Each fuzzer favors seeds with specific traits, which makes tailored seed selection and prioritization a promising optimization direction. For example, MDPFuzz performs better on seeds whose reward scores are highly sensitive to state perturbations, significantly lifting overall fuzzing efficacy.
3) Matching each seed to its optimal fuzzer further improves performance. Researchers can characterize seed preferences of every fuzzer via clustering analysis and build dedicated classifiers to assign suitable fuzzers for individual seeds automatically.

\textbf{Adopt Multi-Level Diversity Metrics.}
Our results reveal that different diversity metrics capture distinct aspects of exploration. 
For example, SeqDivFuzz achieves high input diversity but low output diversity, while QDFuzz shows the opposite on BipedalWalker.
These discrepancies indicate that diverse inputs do not necessarily induce diverse agent behaviors.
Researchers should evaluate diversity from multiple perspectives (e.g., seeds, trajectories) rather than relying on a single metric, and future work should investigate which diversity notions most closely correlate with practical safety benefits~\cite{bartlett2025pursuit}.

\textbf{Use Crash Data Judiciously for RL Repair.}
Although crash-triggering test inputs can improve robustness, not all detected crashes are equally beneficial for repair. We observe that incorporating highly abnormal or out-of-distribution crashes, such as those generated by CureFuzz, can even reduce robustness against other fuzzers, such as Random Testing, due to distribution or contribution shifts during fine-tuning~\cite{DBLP:journals/tosem/YouWCCSLD25}, as shown in Table~\ref{table:cross_fuzzing_evaluation}. In addition, fine-tuning may still introduce regression faults (Table~\ref{table:retraining_impact}). Therefore, practitioners should carefully curate repair data by prioritizing representative and informative crashes while filtering noisy or extreme samples. Repair pipelines should also explicitly monitor regression faults throughout the repair process to prevent unintended performance degradation~\cite{DBLP:conf/sigsoft/You25}. These findings also motivate future research on more effective repair strategies that can better exploit crash data while reducing regressions.


\textbf{Establish Standardized and Fair Evaluation Protocols.}
Our findings affirm that inconsistent baselines, configurations, and benchmarks across studies hinder reliable comparison.
Thus, we encourage researchers to consider a standardized evaluation protocol when developing new RL testing methods.
This includes: 1) comparing against strong and updated baselines (e.g., MDPFuzz and random testing). Random testing achieves competitive performance yet is frequently omitted in prior research, which deserves special attention.
2) testing on tasks of varying complexity, including realistic scenarios (e.g., CARLA), and
3) assessing not only crash counts but also practical utility.

\textbf{Design Scalable and Generalizable RL Testing Fuzzers.} 
Many existing RL fuzzing methods~\cite{mdpfuzz,mdpfuzznew, QDfuzz,seqfuzz,curefuzz,g-model} are tightly coupled to specific environments, handcrafted features, or task-dependent assumptions, resulting in high adaptation costs when migrating to new scenarios. Differences in observation spaces, action formats, and crash oracles often require substantial redesign, retraining, or manual tuning, limiting real-world applicability. 
Moreover, certain design choices further hinder scalability and robustness. For example, overly strict seed pruning (e.g., CureFuzz and SeqDivFuzz) can prematurely restrict exploration, sensitivity-driven guidance (e.g., MDPFuzz) may lead to local optima, manually engineered behavior descriptors (e.g., QDFuzz) reduce generalization, and heavy generative backbones (e.g., diffusion models in G-Model) introduce significant overhead in high-dimensional settings.
Future fuzzing methods should prioritize lightweight, adaptive, and task-agnostic designs. Promising directions include dynamic seed management, diversity-aware guidance to avoid local optima, automatically learned behavior representations instead of handcrafted descriptors, and more scalable generative architectures.

\textbf{Evaluate Under Realistic Deployment Scenarios.}
When designing and evaluating RL fuzzing methods, researchers should prioritize realistic deployment scenarios rather than relying solely on simplified or toy benchmarks. Many existing methods~\cite{mdpfuzz,mdpfuzznew, QDfuzz,seqfuzz,curefuzz,g-model} are primarily evaluated on small-scale or simulated environments, which may not adequately reflect the complexity and constraints of real-world applications. 
In practice, RL agents operate in dynamic and uncertain environments, where testing must satisfy stricter requirements on efficiency, resource consumption, and the practical relevance of detected crashes.
For example, autonomous driving systems must respond to real-time environmental changes and ensure that discovered crashes correspond to meaningful safety risks or functional requirements (e.g., lane-changing or overtaking behaviors)~\cite{DBLP:journals/ese/GiamatteiBPRT25}.
Evaluations that overlook these constraints may overestimate the practical effectiveness of fuzzing methods.

\section{Threats to Validity}
\label{subsec:threats}

\noindent
\textbf{\textit{Internal threat.}} First, implementation bias may affect the results. To mitigate this threat, we adopt the official replication packages provided by the original authors whenever available, and avoid modifying their core logic beyond necessary integration.
Second, differences in computational resources and runtime environments could affect efficiency measurements. We therefore conduct all experiments using the same hardware platform under identical time budgets for fair comparisons.

\smallskip
\noindent
\textbf{\textit{External threat.}} 
First, to mitigate threats arising from biased benchmark selection, we conduct experiments on three representative RL benchmarks that cover increasing levels of complexity, ranging from low-dimensional control tasks to high-dimensional autonomous driving scenarios.
Second, while we include five state-of-the-art fuzzing methods alongside random testing and employ their officially updated implementations to ensure fair comparison, other existing or newly proposed fuzzers may still demonstrate distinct behaviors. To address this threat, we carefully select the evaluated methods based on their representativeness in the literature and the availability of reproducible artifacts.
Finally, the inherent randomness of fuzz testing can introduce variability in experimental results. To mitigate this effect, we repeat each experiment five times and report the average results.


\smallskip
\noindent
\textbf{\textit{Construct threat.}}
Our study operationalizes fuzzing performance using several metrics across effectiveness, diversity, efficiency, and usefulness.
While these metrics are widely adopted in prior work~\cite{mdpfuzz,mdpfuzznew, QDfuzz,seqfuzz,curefuzz,g-model} and provide measurable and comparable indicators, they inevitably approximate the underlying constructs rather than capturing them perfectly. 
To mitigate these threats, we adopt commonly used definitions, evaluate all methods under unified configurations, and assess multiple complementary metrics to provide a more comprehensive and balanced evaluation.
\section{Conclusion}
\label{sec:conclusion}
This work presents a comprehensive empirical study that systematically investigates the effectiveness, diversity, efficiency, and practical utility of RL fuzzing methods. We benchmark five state-of-the-art methods alongside random testing across three representative environments with varying complexity.
Our evaluation yields several key findings. 
MDPFuzz consistently achieves the highest crash detection effectiveness and efficiency, while SeqDivFuzz demonstrates stronger performance in uncovering diverse crash behaviors.
Additionally, we further show that fuzzing-generated test cases provide tangible downstream benefits: they can improve agent robustness by up to 41.5\% through targeted repair and enable safety monitoring agents that generalize effectively across different fuzzing strategies. These results highlight that RL fuzzing is not only useful for exposing crashes but also valuable for enhancing agent reliability and operational safety.

\section{Acknowledgement}
\label{sec:Acknowledgement}
This work was supported by the National Natural Science Foundation of China (Grant Nos. 62322208, 62232001)

\bibliographystyle{IEEEtranS}
\bibliography{main}

\end{document}